\pdfoutput=1
\documentclass[11pt]{article}

\usepackage{acl}

\usepackage{times}
\usepackage{latexsym}
\usepackage{graphicx}
\usepackage{amssymb,amsthm,amsfonts}
\usepackage{amsmath}
\usepackage{hyperref}
\usepackage{booktabs}
\usepackage{multirow}
\usepackage{float}
\usepackage{enumitem}
\usepackage[normalem]{ulem}
\usepackage{subfigure}
\usepackage{bbding}

\usepackage[T1]{fontenc}
\usepackage[utf8]{inputenc}

\usepackage{microtype}

\usepackage{inconsolata}

\title{\ourmethod{}: Robust Spoiler Detection with Multi-modal Information and Domain-aware Mixture-of-Experts}
\author{Zinan Zeng\textsuperscript{1} \ \ \ Sen Ye \textsuperscript{1} \ \ \ Zijian Cai\textsuperscript{1} \ \ \ Heng Wang\textsuperscript{1} \\ \textbf{Yuhan Liu\textsuperscript{1}} \ \ \textbf{Haokai Zhang\textsuperscript{1}} \ \ \ \textbf{Minnan Luo \textsuperscript{1}} \thanks{$^*$Corresponding author: Minnan Luo, School of Computer Science and Technology, Xi’an Jiaotong University, Xi’an 710049, China.}\\
\textsuperscript{1}Xi'an Jiaotong University \\
\small \texttt{\{2194214554, ys2003, 2205114706, wh2213210554, lyh6560, zhanghaokai\}@stu.xjtu.edu.cn} \\  
\small \texttt{minnluo@xjtu.edu.cn}
}
\newcommand{\ourmethod}[1]{}
\renewcommand{\ourmethod}[1]{\texttt{MMoE}}
  
\begin{document}
\maketitle
% \vspace {1cm}
\begin{abstract}
Online movie review websites are valuable for information and discussion about movies. However, the massive spoiler reviews detract from the movie-watching experience, making spoiler detection an important task. Previous methods simply focus on reviews' text content, ignoring the heterogeneity of information in the platform. For instance, the metadata and the corresponding user's information of a review could be helpful. Besides, the spoiler language of movie reviews tends to be genre-specific, thus posing a domain generalization challenge for existing methods. To this end, we propose \ourmethod{}, a multi-modal network that utilizes information from multiple modalities to facilitate robust spoiler detection and adopts Mixture-of-Experts to enhance domain generalization.
\ourmethod{} first extracts graph, text, and meta feature from the user-movie network, the review's textual content, and the review's metadata respectively. To handle genre-specific spoilers, we then adopt Mixture-of-Experts architecture to process information in three modalities to promote robustness. Finally, we use an expert fusion layer to integrate the features from different perspectives and make predictions based on the fused embedding. Experiments demonstrate that \ourmethod{} achieves state-of-the-art performance on two widely-used spoiler detection datasets, surpassing previous SOTA methods by 2.56\% and 8.41\% in terms of accuracy and F1-score. Further experiments also demonstrate \ourmethod{}'s superiority in robustness and generalization. Our code is available at https://github.com/zzqbjt/Spoiler-Detection.

\end{abstract}

\section{Introduction}
Movie websites such as IMDb and Rotten Tomato have served as popular social platforms facilitating commentary, discussion, and recommendation about movies \cite{cao2019unifying}.  
However, there are a substantial amount of reviews that reveal the critical plot in advance on these websites, known as \textit{spoilers}. Spoilers diminish the suspense and surprise of the movie and may evoke negative emotions in the users \cite{loewenstein1994psychology}.
Therefore, it is necessary to propose an effective spoiler detection method to protect users' experience. 

\begin{figure}[t]
\centering
    \includegraphics[width=0.5\textwidth]{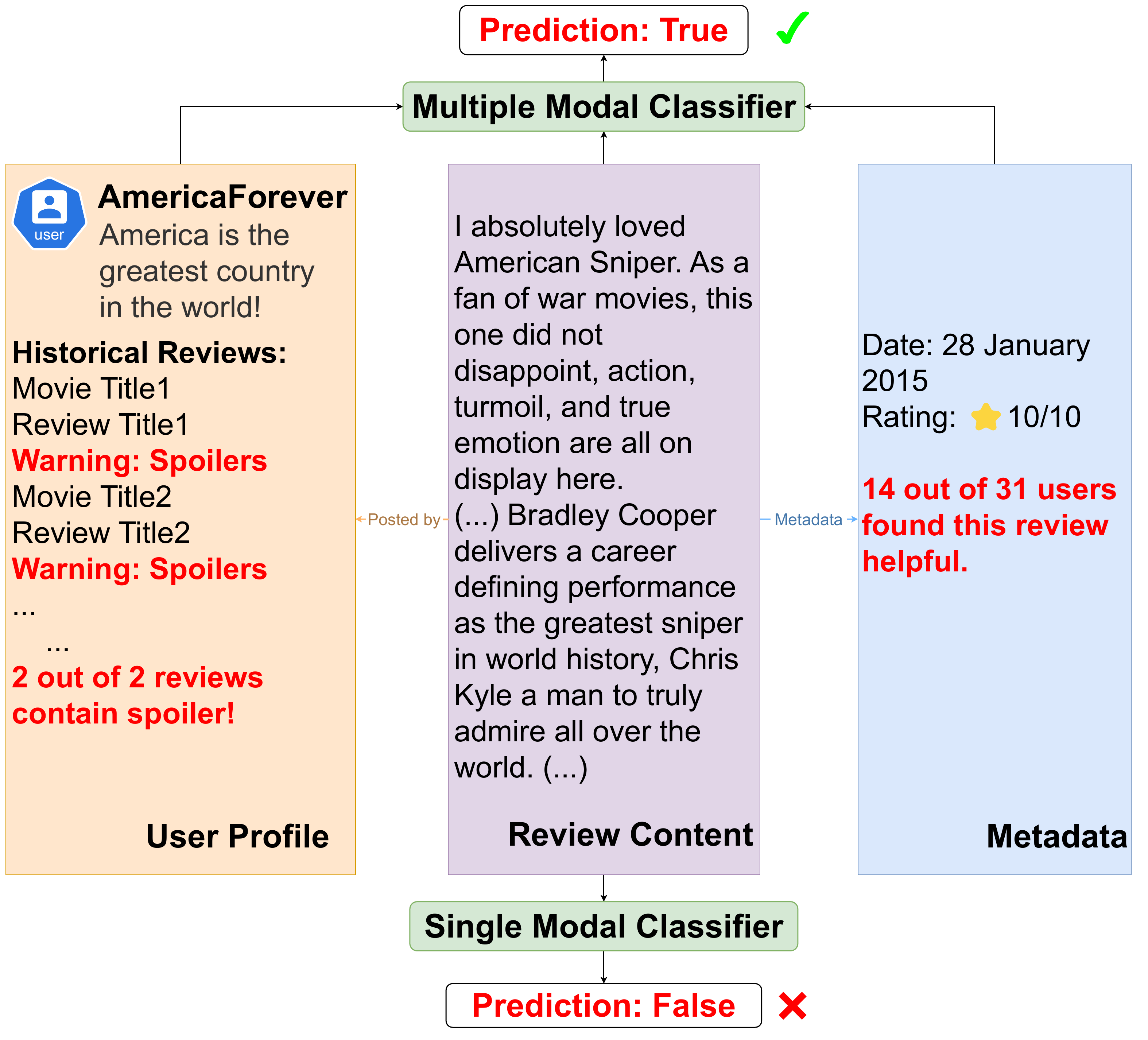}
    \caption{The information of a spoiler review from multiple sources. The text-based detection method struggles to identify whether this review is a spoiler. However, we can identify the review to be a spoiler by jointly considering the reviewer's historical preference and the review's metadata. The red font indicates the information which helps determine whether the review contains spoilers.}
    \label{fig:teaser}
\end{figure}

Existing spoiler detection methods mainly focus on the textual content. \citet{chang2018deep} encode review sentences and movie genres together to detect spoilers. \citet{wan2019fine} incorporate Hierarchical Attention Network \cite{yang2016hierarchical} and introduce user bias and item bias. \citet{chang2021killing} exploit syntax-aware graph neural networks to model dependency relations in context words. \citet{wang2023detecting} take into account external movie knowledge and user interactions to promote effective spoiler detection. 

However, there are still some limitations in the proposed approaches so far. Firstly, solely relying on the textual content is inadequate for robust spoiler detection \cite{wang2023detecting}. We argue that integrating multiple information sources (metadata, user profile, movie synopsis et al.) is necessary for reliable spoiler detection. 
For instance, as shown in Figure \ref{fig:teaser}, it is challenging to discern whether this review contains spoilers solely based on its textual content. However, this reviewer can be correctly identified as a spoiler through the analysis of historical reviews and the establishment of a user profile for this reviewer. In addition, the vote count in metadata also suggests that the review is a potential spoiler.
Secondly, the spoiler language tends to be genre-specific as people's focus varies depending on the genre of movies, resulting in distinct characteristics in their reviews. Specifically, for science fiction films, individuals tend to focus on the quality of special effects. In the case of action movies, the fight scenes become the primary highlight. On the other hand, for suspense movies, the plot takes precedence. Consequently, there is a significant variation of the spoilers in reviews across different domains. Existing methods fail to differentiate these reviews with varying styles, posing challenges in adapting to the increasingly diverse landscape of spoiler reviews.

To address these challenges, we propose \ourmethod{} (Multi-modal Mixture-of-Experts), 
which leverages multi-modal information and domain-aware Mixture-of-Experts. Specifically, we start training multiple encoders for different types of information by using a series of pretext tasks. Next, we use these models to obtain the features of reviews from graph view, text view, and meta view. We then adopt Mixture-of-Experts (MoE) to assign the information from different aspects to certain domains. Finally, we use a transformer encoder to combine the information from all three perspectives.
Experiments demonstrate that \ourmethod{} achieves state-of-the-art performance on two widely-used spoiler detection datasets, surpassing previous SOTA methods by 2.56\% and 8.41\% in terms of accuracy and F1-score. Further extensive experiments also validate our design choices.

\begin{figure*}[t]
    \centering
    \includegraphics[width=\textwidth]{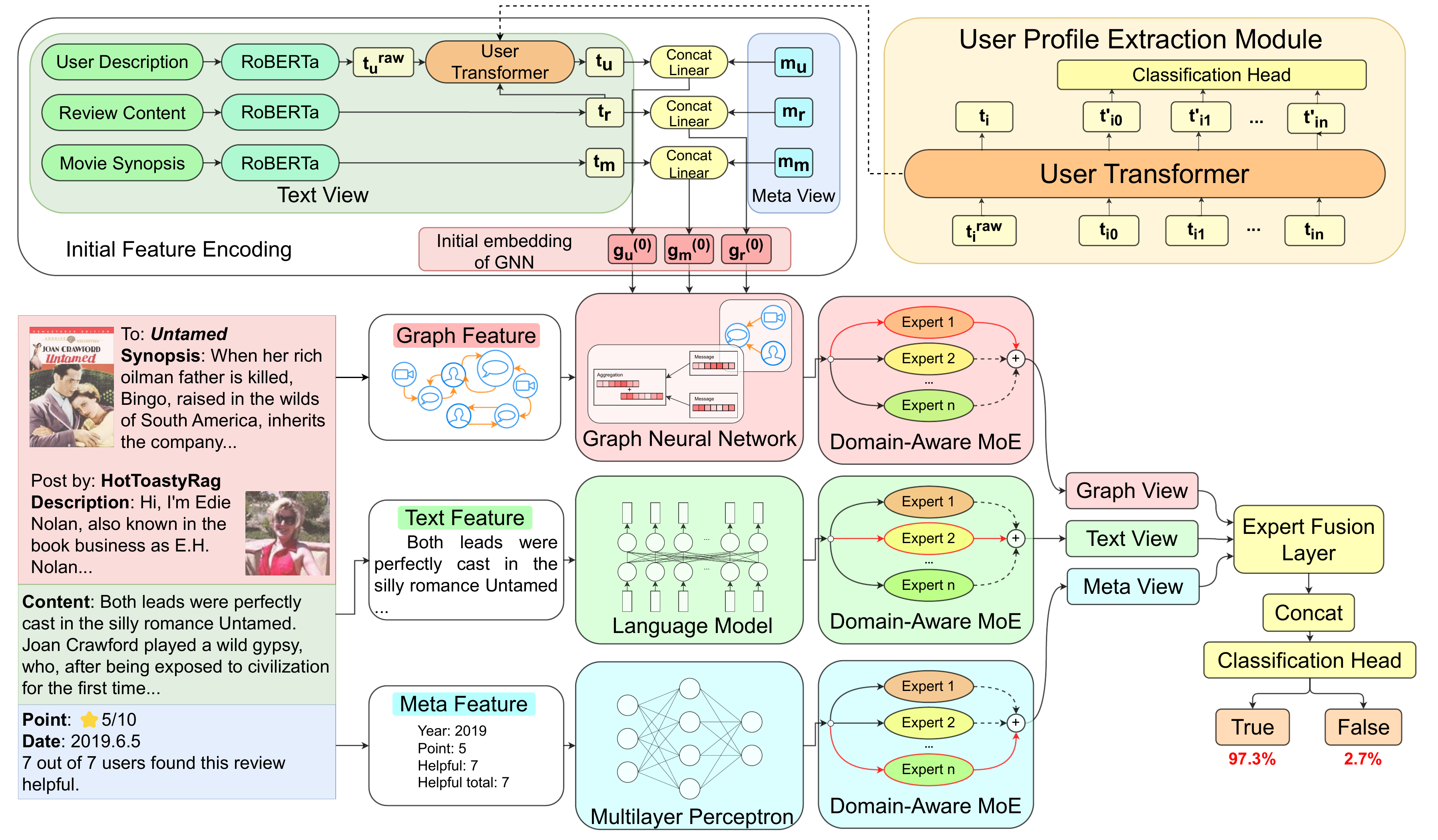}
    \caption{
    \ourmethod{}: a multi-modal mixture-of-experts framework that jointly leverages the review's metadata, text, and graph features for robust and generalizable spoiler detection. Metadata, text, and graph information are first processed by modal-specific encoders and then fed into Mixture-of-Experts layers. The user profile extraction module is employed to analyze the reviewer's historical preference and learn an embedding for the user. Finally, an expert fusion layer is adopted to integrate the three information sources and classify spoilers.
    }
    \label{fig:overview}
    \hfill
\end{figure*}

\section{Related Work}
Spoiler detection aims to automatically detect spoiler reviews in television \cite{boyd2013spoiler}, books \cite{wan2019fine}, and movies \cite{wang2023detecting}, thereby protecting users' experiences. Earlier methods usually design handcrafted features and apply a traditional classifier. \citet{guo2010finding} use bag-of-words embeddings and LDA model \cite{blei2003latent} to detect spoilers in movie comments. \citet{boyd2013spoiler} combine lexical features with metadata features and use an SVM model \cite{cortes1995support} as the classifier. Recently, deep learning based detection methods have dominated. \citet{chang2018deep} propose a model with a genre-aware attention mechanism. However, they don't take into account fine-grained movie text information. \citet{wan2019fine} develop SpoilerNet which uses HAN (Hierarchical Attention Network) \cite{yang2016hierarchical} to learn sentence embeddings and then applies GRU \cite{cho2014learning} on top of it. SpoilerNet also considers user bias and item bias. However, they simply model them as learnable vectors. \citet{chang2021killing} use bi-directional LSTM \cite{hochreiter1997long} to extract word features and feed the embedding into graph neural network to pass and aggregate messages on the dependency graph. However, it is worth noting that the authors only incorporate the movie's genre information at the final pooling stage. These methods basically use RNN-based networks (such as LSTM and GRU) as text encoders, and review contents are the primary or even the only reference information. \citet{wang2023detecting} first introduce user network and external movie knowledge into spoiler detection task and validate its effectiveness. However, their approach falls short of adequately leveraging user information and adopts a simplistic encoding strategy for the text, relying solely on average pooling.

Given the limitations of the above work, we develop a comprehensive framework which leverages multi-modal information and the domain-aware Mixture-of-Experts for robust and generalizable spoiler detection. Our method \ourmethod{} establishes a new state-of-the-art in spoiler detection.

\section{Methodology} 
The overall architecture of  \ourmethod{} is illustrated in Figure \ref{fig:overview}. 
Specifically, we first encode the review's meta, text, and graph information to obtain comprehensive representations from three perspectives. We also propose a user profile extraction module which learns from the reviewer's historical reviews and analyzes the reviewer's preference.
To deal with genre-specific spoilers, we then adopt Mixture-of-Expert (MoE) architecture \cite{jacobs1991adaptive, shazeer2017outrageously} to process features in different modalities. MoE is able to assign reviews with different characteristics to different experts for robust classification \cite{liu2023botmoe}. To facilitate information interaction, we finally use an expert fusion layer to integrate the information from the three perspectives and classify whether the review is a spoiler.

\subsection{Modal-specific Feature Encoder} \label{3.1}
\noindent\textbf{Metadata Encoder.} The metadata associated with spoiler reviews tends to differ from that of regular reviews. Consequently, we gather the review metadata as auxiliary information for classification.  Details of metadata are illustrated in Appendix \ref{A}. Once this numerical information is collected, we employ a two-layer MLP as the meta encoder.

\noindent\textbf{Text Encoder.} 
The textual content plays a crucial role in spoiler detection. To obtain high-quality embeddings, we employ RoBERTa \cite{liu2019roberta} as our text encoder. Initially, we fine-tune RoBERTa through a binary classification task using the textual content of reviews, which ensures that the model is specifically tailored for our spoiler detection task. Subsequently, we utilize the fine-tuned RoBERTa to encode the review content and transform the encoded embedding with a single-layer MLP.  

\noindent\textbf{Graph Encoder.} To model the complex relations and interactions between user, review, and movie, we employ graph neural network to update the review feature through the corresponding user feature and movie feature. We first construct a directed graph consisting of the following three types of nodes and three types of edges: \\
\uline{N0: \emph{User}}. \\
\uline{N1: \emph{Movie}}. \\
\uline{N2: \emph{Reviews}}. \\
\uline{E1: \emph{Movie-Review}} We connect a review node with a movie node if the review is about the movie. \\
\uline{E2: \emph{User-Review}} We connect a review node with a user node if the user posts the review. \\
\uline{E3: \emph{Review-User}} We use this type of edge to enable message passing between reviews.

For movie and review nodes, we encode their synopsis and review content respectively by the fine-tuned RoBERTa as the input feature. For user nodes, we design a user profile extraction module (Section \ref{user profile extraction module}) to extract their profiles as the initial feature. Initial node features are transformed by a linear layer followed by a ReLU activation, \emph{i.e.},
\begin{align*}
\bold{g}^{(0)}_\textit{i}=\max(\bold{W}_{\textit{in}} \cdot [\bold{t}_\textit{i}, \bold{m}_\textit{i}] + \bold{b}_{\textit{in}}, 0),
%    \nonumber
\end{align*}
where $\bold{m}_\textit{i}$, $\bold{t}_\textit{i}$ and $\bold{g}^{(0)}_\textit{i}$ denote metadata features, text features and the initial embedding in the graph of node $i$. $[\cdot, \cdot]$ denotes the concatenation operation. $\bold{W}_{\textit{in}}$ and $\bold{b}_{\textit{in}}$ are parameters of the linear layer. We then use Graph Attention Network (GAT) \cite{velickovic2017graph} as the graph encoder to obtain the embedding of reviews from the graph modality, \emph{i.e.},

\begin{align*}   
\bold{g}_\textit{i}^{(l+1)}
&=\alpha_{\textit{i,i}}\bold{\Theta}_\textit{s}\bold{g}_\textit{i}^{(l)}+\sum_{j\in N(i)}{\alpha_{\textit{i,j}}\bold{\Theta}_\textit{t}\bold{g}_\textit{j}^{(l)}},
%    \nonumber
\end{align*}
\begin{small}
    \begin{align*}   \alpha_{\textit{i,j}}=\frac{\exp{(f(\bold{a}_\textit{s}^\textit{T}\bold{\Theta}_\textit{s}\bold{g}_\textit{i}^{(l)}+\bold{a}_\textit{t}^\textit{T}\bold{\Theta}_\textit{t}\bold{g}_\textit{j}^{(l)}))}}{\sum_{k\in N(i)\cup{i}}\exp{(f(\bold{a}_\textit{s}^\textit{T}\bold{\Theta}_\textit{s}\bold{g}_\textit{i}^{(l)}+\bold{a}_\textit{t}^\textit{T}\bold{\Theta}_\textit{t}\bold{g}_\textit{j}^{(l)}}))}
%    \nonumber
    \end{align*}
\end{small}where $f$ denotes the Leaky ReLU activation function. $\bold{g}_\textit{i}^{(l)}$ is the embedding of node $i$ in layer {l}. $N(i)$ is the neighbors of node $i$. In the directed graph, $N(i)$ denotes all nodes which point to node $i$. $\alpha_{i,j}$ is the attention score between node $i$ and node $j$. $\bold{\Theta}_\textit{s}\in \mathbb{R}^{d_{\textit{in}}\times d_{\textit{out}}}$, $\bold{\Theta}_\textit{t}\in \mathbb{R}^{d_{\textit{in}}\times d_{\textit{out}}}$, $\bold{a}_\textit{s}\in \mathbb{R}^{d_{\textit{in}}}$, $\bold{a}_\textit{t}\in \mathbb{R}^{d_{\textit{in}}}$ are learnable parameters. 
$d_{\textit{in}}$ and $d_{\textit{out}}$ are the dimension of input vector and output vector, respectively. 

We add a ReLU activation function between every GAT layer. After $L$ layers of GAT, we obtain the review embeddings from the graph view.

\subsection{User Profile Extraction Module} \label{user profile extraction module}
Since users normally have their preferences, they either infrequently or frequently post spoiler reviews. The specific proportion of spoiler reviews per user can be found in Appendix \ref{A}, which illustrates this bias in detail. Therefore, capturing user preferences through their profiles can significantly aid in spoiler detection. While using users' self-descriptions is a direct approach to obtain their profiles, unluckily most users do not provide descriptions on film websites. Therefore, the initial information of user nodes is often missing in the graph. In light of this challenge, we model this kind of user preference by obtaining a learned user profile embedding through a user profile extraction module which takes the user's historical reviews as input and outputs a summarizing embedding indicating the user's preference.

To be specific, we concatenate the raw semantic features of users and the semantic features of their reviews into a sequence, \emph{i.e.},
\begin{align}
    \bold{s}_\textit{i}=[\bold{t}_\textit{i}^{\textit{raw}}, \bold{t}_{\textit{i}_1}, \bold{t}_{\textit{i}_2}, \cdots ,\bold{t}_{\textit{i}_\textit{n}}]
    \nonumber
\end{align}
where $\bold{t}_\textit{i}^{\textit{raw}}$ is the raw text feature of the $i$-th user's description encoded by RoBERTa, $\bold{t}_{\textit{i}_1}$, $\bold{t}_{\textit{i}_2}, \cdots,\bold{t}_{\textit{i}_\textit{n}}$ are the text feature of the first, second, $\cdots$ and the last review of user $i$. $\bold{s}_\textit{i}$ is the input sequence of the module. 
Since the number of reviews per user can vary,  we employ the ``maximum length'' strategy. Sequences shorter than the maximum length are padded with zero vectors, while sequences longer than the maximum length are truncated to ensure uniform length.

After obtaining the input sequence, we use a transformer encoder \cite{vaswani2017attention} to get the output sequence. 
The encoder summarizes the user's historical reviews and utilizes self-attention mechanisms to learn a comprehensive profile embedding that reflects the user's preference. We pre-train the encoder by attaching a classification head after each review embedding, \emph{i.e.},
\begin{align*}
    \bold{s}'_\textit{i}&=\mathrm{TRM}(\bold{s}_\textit{i}),\\
    \hat{\bold{p}}_\textit{i}&=\mathrm{softmax}(\bold{W}_\textit{u} \cdot \bold{s}'_\textit{i}+\bold{b}_\textit{u}),
\end{align*}
where $\bold{s}'_\textit{i}$ is the output sequence; $\hat{\bold{p}}_\textit{i}$ is the predicted output. We only compute the loss for the reviews within the training set.

After pre-training, we use the encoder to perform forward propagation on all sequences and extract the first embedding in the sequence (corresponding to the position of the user’s raw profile feature in the input) as the user’s profile feature, denoted as $\bold{t}_\textit{i}$.
The embedding will then be fixed in the model by
\begin{align*}
    \bold{t}_\textit{i} = \bold{s}'_\textit{i}[0].
\end{align*}

\subsection{Domain-Aware MoE Layer} \label{moe layer}
Inspired by the successful applications of Mixture-of-Experts in NLP and bot detection \cite{shazeer2017outrageously, fedus2022switch, liu2023botmoe}, we adopt MoE to divide and conquer the information in the three modalities. Since spoiler reviews exhibit distinct characteristics across different genres of movies, we leverage the MoE framework, activating different experts to handle different reviews belonging to various domains. We calculate the weight $G_\textit{j}$ of each expert $E_\textit{j}$ as the same as \citet{shazeer2017outrageously}. Each expert $E_\textit{j}$ is a 2-layer MLP, \emph{i.e.},
\begin{align}
    \bold{z}^{\textit{mod}}_\textit{i}=\sum_{j=1}^{n}{G_\textit{j}(\bold{x}^{\textit{mod}}_\textit{i})E_\textit{j}(\bold{x}^{\textit{mod}}_\textit{i})},
    \nonumber
\end{align}
where $\bold{x}^{\textit{mod}}_\textit{i}$ is the input embedding of review $i$, $\bold{z}^{\textit{mod}}$ is the output feature, and $\textit{mod} \in \{ m, t, g \}$. 

\begin{table*}[h]
\centering
\caption{Accuracy, AUC, and binary F1-score of \ourmethod{} and other baselines on the two datasets. We repeat all experiments five times and report the average performance with standard deviation. \textbf{Bold} indicates the best performance, \uline{underline}
the second best. \ourmethod{} significantly outperforms the previous state-of-the-art method on two benchmarks on all metrics.}
\resizebox{\textwidth}{!}
{
\begin{tabular}{lcccccc}
\toprule
\multirow{2}{*}{\textbf{Model}}
& \multicolumn{3}{c}{\textbf{Kaggle}}
& \multicolumn{3}{c}{\textbf{LCS}}\\
\cmidrule{2-4}
\cmidrule{5-7}
& \textbf{F1} & \textbf{AUC} & \textbf{Acc} 
& \textbf{F1} & \textbf{AUC} & \textbf{Acc}\\
\midrule
BERT \cite{devlin2018bert} 
& 44.02 ($\pm$1.09) & 63.46 ($\pm$0.46) & 77.78 ($\pm$0.09)
& 46.14 ($\pm$2.84) & 65.55 ($\pm$1.36) & 79.96 ($\pm$0.38)\\
RoBERTa \cite{liu2019roberta} 
& 50.93 ($\pm$0.76) & 66.94 ($\pm$0.40) & 79.12 ($\pm$0.10) 
& 47.72 ($\pm$0.44) & 65.55 ($\pm$0.22) & 80.16 ($\pm$0.03) \\
BART \cite{lewis2019bart} 
& 46.89 ($\pm$1.55) & 64.88 ($\pm$0.71) & 78.47 ($\pm$0.06) 
& 48.18 ($\pm$1.22) & 65.79 ($\pm$0.62) & 80.14 ($\pm$0.07) \\
DeBERTa \cite{he2021debertav3} 
& 49.94 ($\pm$1.13) & 66.42 ($\pm$0.59) & 79.08 ($\pm$0.09) 
& 47.38 ($\pm$2.22) & 65.42 ($\pm$1.08) & 80.13 ($\pm$0.08) \\
\midrule
GCN \cite{kipf2016semi} 
& 59.22 ($\pm$1.18) & 71.61 ($\pm$0.74) & 82.08 ($\pm$0.26) 
& 62.12 ($\pm$1.18) & 73.72 ($\pm$0.89) & 83.92 ($\pm$0.23) \\
R-GCN \cite{schlichtkrull2018modeling} 
& 63.07 ($\pm$0.81) & 74.09 ($\pm$0.60) & 82.96 ($\pm$0.09) 
& 66.00 ($\pm$0.99) & 76.18 ($\pm$0.72) & 85.19 ($\pm$0.21) \\
GAT \cite{velickovic2017graph}
& 60.98 ($\pm$0.09) & 72.72 ($\pm$0.06) & 82.43 ($\pm$0.01) 
& 65.73 ($\pm$0.12) & 75.92 ($\pm$0.13) & 85.18 ($\pm$0.02) \\
SimpleHGN \cite{lv2021we}
& 60.12 ($\pm$1.04) & 71.61 ($\pm$0.74) & 82.08 ($\pm$0.26) 
& 63.79 ($\pm$0.88) & 74.64 ($\pm$0.64) & 84.66 ($\pm$1.61) \\
\midrule
DNSD \cite{chang2018deep} 
& 46.33 ($\pm$2.37) & 64.50 ($\pm$1.11) & 78.44 ($\pm$0.12) 
& 44.69 ($\pm$1.63) & 64.10 ($\pm$0.74) & 79.76 ($\pm$0.08) \\
SpoilerNet \cite{wan2019fine} 
& 57.19 ($\pm$0.66) & 70.64 ($\pm$0.44) & 79.85 ($\pm$0.12) 
& 62.86 ($\pm$0.38) & 74.62 ($\pm$0.09) & 83.23 ($\pm$1.63) \\
MVSD \cite{wang2023detecting} 
& \uline{65.08} ($\pm$0.69) & \uline{75.42} ($\pm$0.56) & \uline{83.59} ($\pm$0.11) 
& \uline{69.22} ($\pm$0.61) & \uline{78.26} ($\pm$0.63) & \uline{86.37} ($\pm$0.08) \\
\midrule
\ourmethod{} (Ours)
& \textbf{71.24} ($\pm$0.08) & \textbf{79.61} ($\pm$0.09) & \textbf{86.00} ($\pm$0.04) 
& \textbf{75.04} ($\pm$0.06) & \textbf{82.23} ($\pm$0.04) & \textbf{88.58} ($\pm$0.02) \\
\bottomrule
\end{tabular}
}
\label{result}
\end{table*}

\subsection{Expert Fusion Layer} \label{fusion layer}
After obtaining the review's representations processed by domain-aware experts in three modalities, we further combine the representations in three modalities by a multi-head transformer encoder to facilitate modality interaction, \emph{i.e.},
\begin{align*}
    \bold{u}_\textit{i}&=[\bold{z}^\textit{m}_\textit{i},\bold{z}^\textit{t}_\textit{i},\bold{z}^\textit{g}_\textit{i}],\\
    \bold{v}_\textit{i}&=\mathrm{TRM}(\bold{u}_\textit{i}),
\end{align*}
where $\bold{z}^\textit{m}_\textit{i}$, $\bold{z}^\textit{t}_\textit{i}$, $\bold{z}^\textit{g}_\textit{i}$ are features from the meta view, text view, and graph view respectively. $\bold{u}_\textit{i}$ represents the concatenated sequence and $\bold{v}_\textit{i}$ denotes the output sequence by the transformer encoder. We finally flatten $\bold{v}_\textit{i}$ and apply a linear output layer to classify, \emph{i.e.},
\begin{align}
    \hat{\bold{y}}_\textit{i}=\bold{W}_\textit{o} \cdot \mathrm{flatten}(\bold{v}_\textit{i})+\bold{b}_\textit{o}.
    \nonumber
\end{align}

\subsection{Learning and Optimization} \label{learn}
We optimize the network by cross-entropy loss with $L_2$ regularization and balancing loss. The total loss function is as follows:

\begin{small}
    \begin{align}
        Loss=\ -\sum{\bold{y}_\textit{i}\log{{\hat{\bold{y}}}_\textit{i}}}+\lambda\sum\theta^2+w\sum_{\textit{mod}}^{m,t,g} B L(\bold{x}^{\textit{mod}}_{\textit{i}}),
        \nonumber
    \end{align}
\end{small}

where ${\hat{\bold{y}}}_\textit{i}$ and $\bold{y}_\textit{i}$ are the prediction for $i$-th review and its corresponding ground truth, respectively. $\theta$ denotes all trainable model parameters, and $\lambda$ and $w$ are hyperparameters which maintain the balance among the three parts. For balancing loss $B L(\bold{x}) = C V (\sum_i G(\bold{x}_\textit{i}))^2$, where $CV$ denotes the coefficient of variation, $G(\bold{x}_\textit{i})$ denotes the calculated weight of each expert, we refer to \citet{shazeer2017outrageously} to encourage each expert to receive a balanced sample of reviews.

\section{Experiment}
\subsection{Experiment Settings}
\textbf{Dataset.}  We evaluate our method \ourmethod{} on \textbf{LCS} dataset \cite{wang2023detecting} and
\textbf{Kaggle IMDB Spoiler} dataset \cite {misra2019imdb}. We follow the same dataset split method as \citet{wang2023detecting}. Specific details of datasets can be found in Appendix \ref{A}.

\noindent \textbf{Baselines.}  We use the same baselines as in \citet{wang2023detecting}. Specifically, we explore three kinds of approaches: PLM(Pre-trained Language Model)-based methods, GNN(Graph Neural Network)-based methods, and task-specific methods. For PLM-based methods,  We evaluate BERT \cite{devlin2018bert}, RoBERTa \cite{liu2019roberta}, BART \cite{lewis2019bart} and DeBERTa \cite{he2021debertav3}. For GNN-based methods, we evaluate GCN \cite{kipf2016semi}, R-GCN \cite{schlichtkrull2018modeling}, GAT \cite{velickovic2017graph}, and Simple-HGN \cite{lv2021we}. For task-specific moethods, we evaluate DNSD \cite{chang2018deep}, SpoilerNet \cite{wan2019fine}, and MVSD \cite{wang2023detecting}. Specific details of baselines can be found in Appendix \ref{D}.

\noindent \textbf{Implementation Details.} We use Pytorch \cite{paszke2019pytorch}, Pytorch Geometric \cite{fey2019fast}, scikit-learn \cite{pedregosa2011scikit}, and Transformers \cite{wolf2020transformers} to implement \ourmethod{}. The hyperparameter settings and architecture parameters are shown in Appendix \ref{B}. We conduct our experiments on a cluster with 4 Tesla V100 GPUs with 32 GB memory, 16 CPU cores, and 377GB CPU memory.

\begin{figure*}[t]
    \centering
    \subfigure[Removing rate of graph edges]{
        \includegraphics[width=0.32\linewidth]{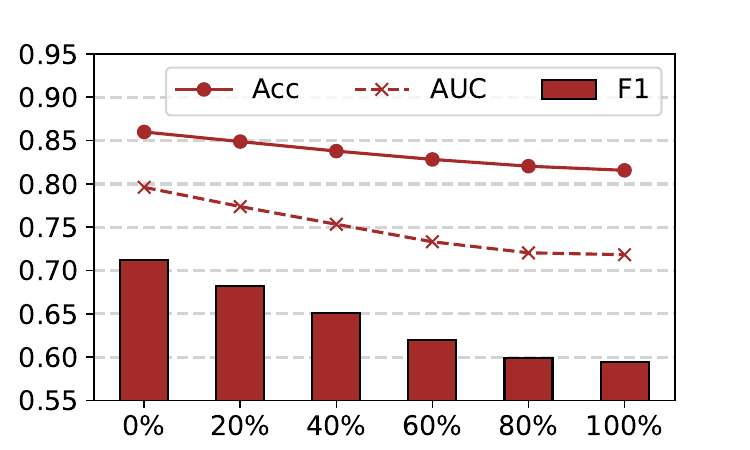}}
    \subfigure[Removing rate of text features]{
        \includegraphics[width=0.32\linewidth]{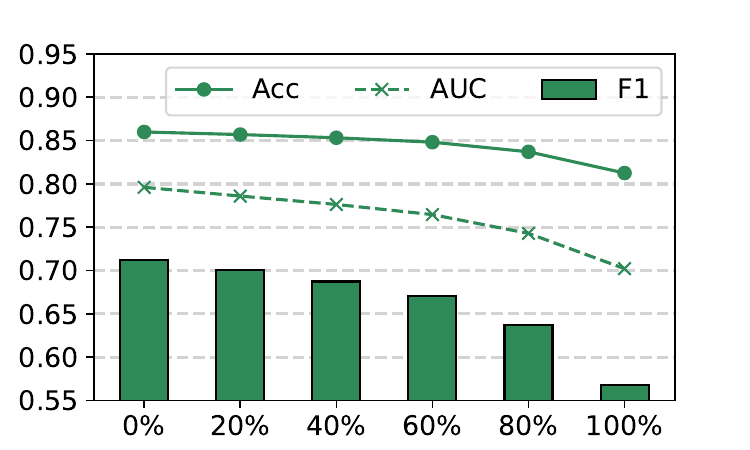}}
    \subfigure[Removing rate of meta features]{
        \includegraphics[width=0.32\linewidth]{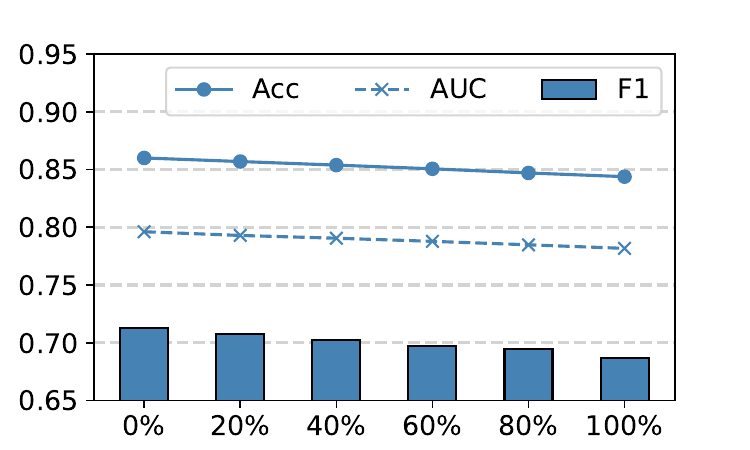}}
    \caption{\ourmethod{} performance when randomly removing edges in the graph, setting elements of text features to zero, and setting elements of meta features to zero. Performance slowly declines with the gradual ablations, indicating the robustness of our method.}
    \label{fig:remove}
\end{figure*}

\subsection{Overall Performances}
We evaluate our proposed \ourmethod{} and other baseline methods on the two datasets. The results presented in Table \ref{result} demonstrate that: 
\begin{itemize}[leftmargin=9pt]
\item \ourmethod{} achieves state-of-the-art on both datasets, outperforming all other methods by at least 8.41\% in F1-score, 5.07\% in AUC, and 2.56\% in accuracy. This illustrates that \ourmethod{} is not only more accurate but also much more robust than former approaches. 
\item GNN-based methods significantly outperform other types of baselines. This confirms our view that using text information alone is not enough in spoiler detection. Social network information from movies and users is also very important.
\item For task-specific baselines, SpoilerNet \cite{wan2019fine} outperforms DNSD \cite{chang2018deep} with user bias. MVSD \cite{wang2023detecting}, which introduces graph neural networks to handle user interactions, undoubtedly performs best. \ourmethod{} further reinforces user bias and thus achieves much better results.
\end{itemize}

\begin{figure}[t]
    \centering
    \subfigure[The attention score between modalities]{
        \includegraphics[width=0.4\linewidth]{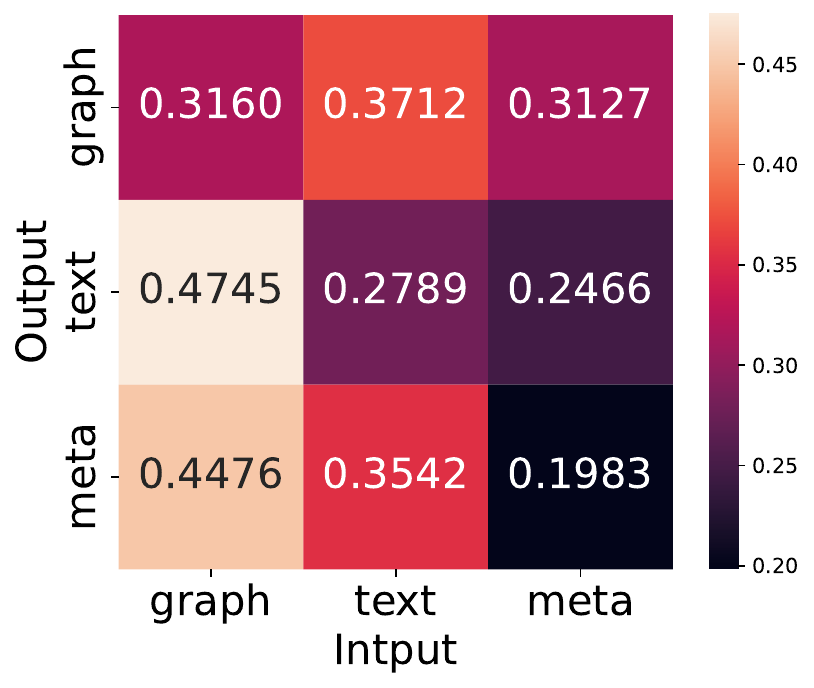}}
    \quad
    \subfigure[The attention score of edges in GNN]{
    \includegraphics[width=0.4\linewidth]{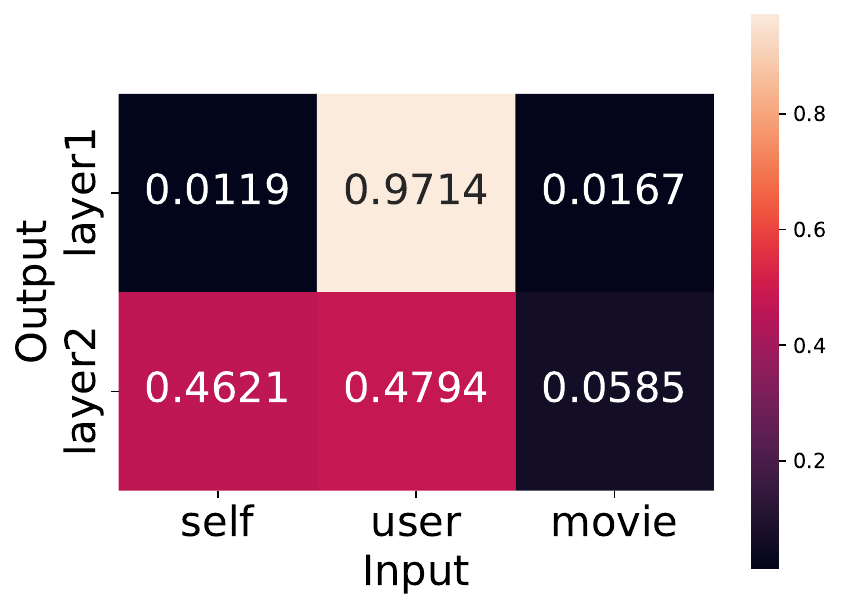}}
    \caption{We first investigate the contribution of information from the three views (graph, text, and meta). We then delve into the graph neural network to find out which nodes the review nodes mainly receive information from.}
    \label{fig:multi-modal}
\end{figure}

\begin{figure}[t]
    \centering
    \includegraphics[width=\linewidth]{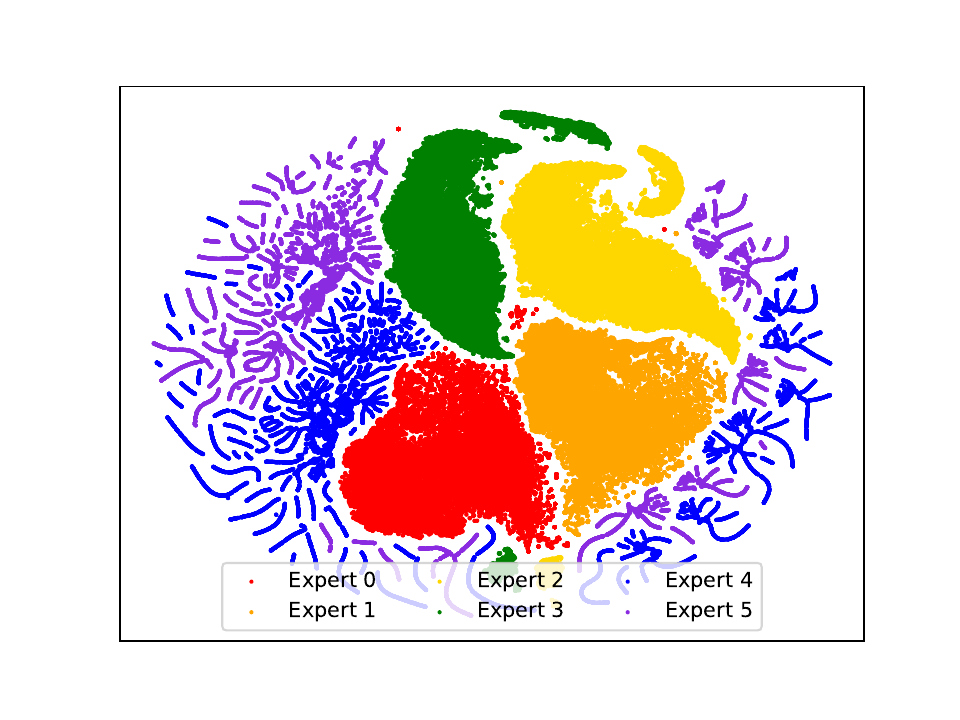}
    \caption{T-SNE visualization of reviews’ graph, text, and meta features. Reviews of the same expert are represented in the same color. The reviews are clearly divided into domains based on their embedding.}
    \label{fig:tsne}
\end{figure}

\subsection{Robustness Study}
We verify the robustness of the model by randomly perturbing the input to simulate the absence of some information reviewed in the real situation. In specific, for graph view information, we randomly remove some of the edges in the graph; for text view and meta view information, we randomly set some of the elements to zero. The result in Figure \ref{fig:remove} shows that, with the help of information from other modalities, even if some of the information is missing, our model still makes the correct prediction most of the time. This proves our view that multi-source information can not only improve the prediction accuracy of the model but also enhance the robustness of the model. 

\subsection{Multi-Modal Study}
To further investigate the contribution of information from each modality, we calculate the attention score between features of different views. In specific, we extract the attention score of each layer in the final expert fusion transformer, and average the score of each layer. Then by averaging the values of each sample, we obtain the heat map as shown in Figure \ref{fig:multi-modal}. Graph view features are without doubt the most contributed information, with an average attention score of 0.4127. For graph view features, we expect review nodes to receive sufficient information from user nodes and movie nodes. So we then extract the average attention scores corresponding to different types of edges in each GAT layer. "Self", "user", and "movie" represent the attention scores between review nodes and themselves, corresponding user nodes, and corresponding movie nodes in each layer respectively. It is clear that users' information is the most helpful, which also demonstrates the importance and effectiveness of our designed user profile extraction module.

\subsection{Review Domain Study}
We posit that due to significant variations in review styles across different types of movies, it is essential to categorize them into distinct domains and assign them to appropriate experts using the Mixture-of-Experts (MoE) approach. To validate our hypothesis, we employ T-SNE visualization \cite{van2008visualizing} to depict the domain assignments of reviews. We extract review representations from the MoE's output for the graph, text, and meta modalities and present them in Figure \ref{fig:tsne}. The visualization clearly illustrates that reviews are distinctly segregated into different domains within each modality, which demonstrates the effectiveness of the MoE in categorizing reviews based on their representations.

\begin{table}[h]
\centering
\caption{Ablation study concerning pretext task, user bias, multi-view data, MoE structure, and fusion methods.}
\resizebox{0.48\textwidth}{!}
{
\begin{tabular}{lcccc}
\toprule
\textbf{Category} & \textbf{Setting} & \textbf{F1} & \textbf{AUC} & \textbf{Acc} \\
\midrule
\textbf{Fine-tuning}
& w/o fine-tuning & 67.45 & 76.99 & 84.48 \\
\midrule
\textbf{User profile} 
& w/o user profile & 68.82 & 77.76 & 85.24 \\
\midrule
\multirow{4}{*}{\textbf{Multi-view}} 
& w/o graph view & 58.29 & 71.09 & 81.55 \\
& w/o text view & 70.69 & 79.19 & 85.76 \\
& w/o meta view & 70.00 & 78.59 & 85.64 \\
& replace GAT with R-GCN & 70.34 & 79.03 & 85.51 \\
\midrule
\multirow{4}{*}{\textbf{MoE}} 
& w/o MoE & 70.99 & 79.35 & 85.96 \\
& replace MoE with MLP & 71.09 & 79.43 & 85.97 \\
& 8 experts & 70.96 & 79.40 & 85.84 \\
& 4 experts & 70.93 & 79.29 & 85.94 \\
\midrule
\multirow{3}{*}{\textbf{Fusion}} 
& concatenate & 70.02 & 78.48 & 85.82 \\
& mean-pooling & 69.84 & 78.31 & 85.82 \\
& max-pooling & 70.65 & 78.99 & 85.96 \\
\midrule
\textbf{Ours} 
& \ourmethod{} & \textbf{71.24} & \textbf{79.61} & \textbf{86.00} \\
\bottomrule
\end{tabular}
}
\label{ablation}
\end{table}

\begin{table*}[t]
\centering
\caption{Examples of the performance of two baselines and \ourmethod{}. Underlined parts indicate the plots. "Key Information" indicates the most helpful information from other sources when detecting spoilers.}
\resizebox{\textwidth}{!}
{
\begin{tabular}{llcccc}
\toprule
\textbf{Review Text} & \textbf{Key Information} & \textbf{Label} 
& \textbf{GAT} & \textbf{RoBERTa} & \textbf{\ourmethod{}} \\
\midrule
\uline{A loser called Brian is born on the same night as Jesus of } 
& \textbf{Movie Synopsis:} Brian Cohen is born in a stable a few   
& \multirow{4}{*}{True}\\
\uline{Nazareth. He lives a parallel life with Jesus of Nazareth. }
& doors down from the one in which Jesus is born, (...) His  
& & False & False & True \\
\uline{He joins 'People's Front of Judea', a Jewish revolutionary }   
& desire for Judith and hatred for the Romans lead him to   
& & \XSolidBrush & \XSolidBrush & \CheckmarkBold \\ 
\uline{party, against Romans and is confused as a messiah by} (...) 
& join the People's Front of Judea (...) \\

\midrule
zzzzz. i fell asleep toward the end of this dull, lackluster
& \textbf{User Profile:} 
& \multirow{4}{*}{True}\\
Hollywood product, so i don't know if it's fair to review
& Historical reviews:
& & False & False & True \\
it but i don't feel like going back and watching the ending
& Reviews 1: \textbf{Warning: Spoilers}    
& & \XSolidBrush & \XSolidBrush & \CheckmarkBold \\ 
because i really don't care what happens to (...) 
& Reviews 2: \textbf{Warning: Spoilers} \\

\midrule
With all due respect to the original Star Wars (which is the 
&  
& \multirow{4}{*}{False}\\
greatest movie of all time), this is a spectacular movie, that
& 
& & True & True & False \\
long after you see it, you still find yourself wondering
&  
& & \XSolidBrush & \XSolidBrush & \CheckmarkBold \\ 
about details. (...)
& \\

\midrule
Hacksaw Ridge is an unflinching, violent assault on your  
&  
& \multirow{4}{*}{False}\\
senses with action sequences and people being blown apart,  
& 
& & True & True & False \\
shot in the head, losing limbs etc etc etc which reminded  
&  
& & \XSolidBrush & \XSolidBrush & \CheckmarkBold \\ 
me of the brutal opening scenes in (...)
& \\

\bottomrule
\end{tabular}
}
\label{examples}
\end{table*}

\subsection{Ablation Study}
In order to investigate the effects of different parts of our model on performance, we conduct a series of ablation experiments on the Kaggle dataset. We report the binary F1-Score, AUC, and accuracy of the ablation study in Table \ref{ablation}. 

\noindent \textbf{Fine-tuning Strategy Study.} We remove the fine-tuning step. As we can see, the performance of the model will be significantly reduced across the board. This indicates that the encoding quality of language models is very important for spoiler detection. 

\noindent \textbf{User Profile Study.}  We remove the additional user profile in our model to examine its contribution. The results show that all aspects of the model performance are reduced after removing the user profile, especially F1 and AUC. 

\noindent \textbf{Multi-view Study.} We examine the contribution of information from different perspectives to the final result by removing information from each modality. The graph view information is the most important, which further demonstrates the significance of external information in spoiler detection. We also replace the GAT layer with other layers to observe the effects of different graph convolution operators. Interestingly, R-GCN, which is the best performer in GNN-based baselines, underperforms GAT when applied in our model. In addition, the removal of meta or text view information also has a considerable impact on the final performance, indicating the importance of the multi-view framework.

\noindent \textbf{MoE Study.} To investigate the contribution of MoE, we analyze the performance changes of the model under the condition of removing the entire MoE layer, replacing MoE with MLP, and changing the number of experts. We can find from the results that the MoE layer enables the model to make a more accurate and robust prediction, which proves that it is helpful to divide reviews into different domains. We further change the number of experts to explore its impact. We use 2 experts as default, then increase the number of experts to 4 and 8. The model performance decreases in both settings, indicating that the number of experts needs to be appropriate.

\noindent \textbf{Fusion Strategy Study.}  Finally, we study the effect of the information fusion method on performance. The results show that our self-attention-based transformer fusion method performs best in all aspects. In addition, the performance of the max-pooling method is significantly better than that of concatenation and mean-pooling. 

\subsection{Case Study}
We conduct qualitative analysis to explore the effect of multiple source information. We select some representative cases as shown in Table \ref{examples}. In the first case, the underlined part reveals the main plot of the movie. However, baseline models mainly focus on the review content itself and don't realize that it contains spoilers. With the help of information from the movie synopsis, \ourmethod{} is able to discriminate that the review is a spoiler. As for the second case, it is actually hard to identify whether the review contains spoilers. Yet through the user profile extraction module we designed, we find that the user often posts spoiler reviews. Therefore, a positive label is assigned to the sample. 

\section{Conclusion}
We propose \ourmethod{}, a state-of-the-art spoiler detection framework which jointly leverages features from multiple modalities and adopts a domain-aware Mixture-of-Experts to handle genre-specific spoiler languages. Extensive experiments illustrate that \ourmethod{} achieves the best result among existing methods, highlighting the advantages of multi-modal information, domain-aware MoE, and user profile modeling.


%\section*{Acknowledgement}
%We are grateful to Zhaoxuan Tan from University of Notre Dame for his valuable advice, corrections and inspiration. We would also like to thank all LUD lab members for fostering a collaborative research environment.

\section*{Limitations and Future Work}
We have considered using large language models (LLMs) to profile users based on their historical comments by generating more interpretive text features of users. However, due to the large number of users in the dataset, either calling the LLM through the API or running the open-source LLM locally takes a long time, which is one of the most difficult problems. In addition, the user descriptions generated by the LLM are not necessarily appropriate for our task. However, we still believe that there is considerable potential for using LLM for data augmentation. We can also look beyond user descriptions. Many movies lack plot synopsis. Using LLM to generate synopsis for these movies is also promising. The application of LLMs may be a key factor in subsequent breakthroughs. 

\section*{Ethical Statements}
Although \ourmethod{} has achieved excellent results, it still needs to be carefully applied in practice. Firstly, there is still room for improvement in the performance of \ourmethod{}. We think it's better suited as a pre-screening tool that needs to be combined with human experts to make final decisions. Secondly, the language model encodes social bias and offensive language in the dataset \cite{li2022herb, nadeem2020stereoset}. In addition, the user profile extraction module we introduced may exacerbate this bias. We look forward to further work to detect and mitigate social bias in the spoiler detection task.

\bibliography{custom}

\clearpage
\appendix
\section{Data Details} \label{A}
\begin{figure*}[t]
    \centering
    \subfigure[]{
        \includegraphics[width=0.3\linewidth]{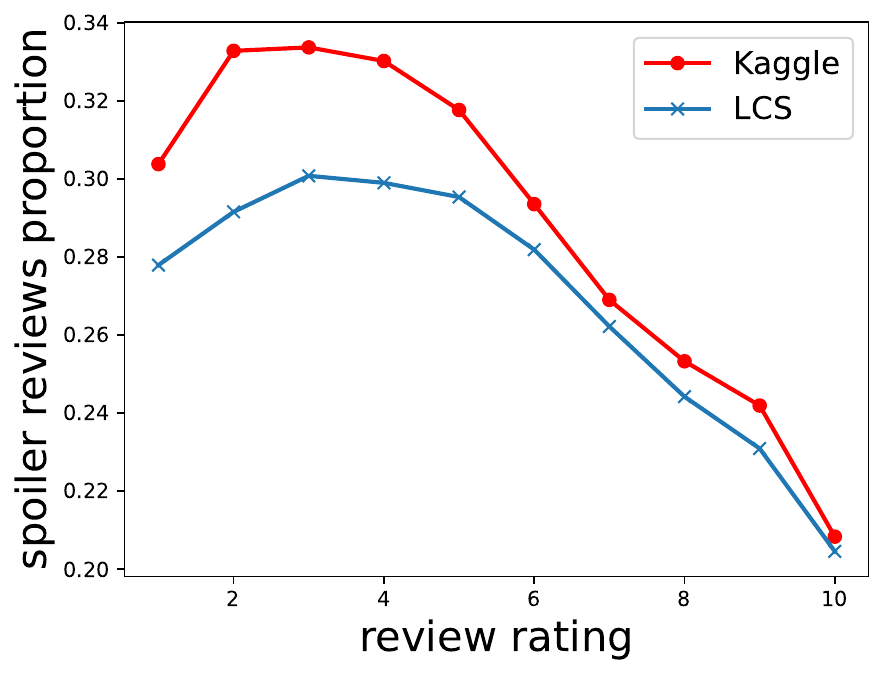}}
    \subfigure[]{
        \includegraphics[width=0.3\linewidth]{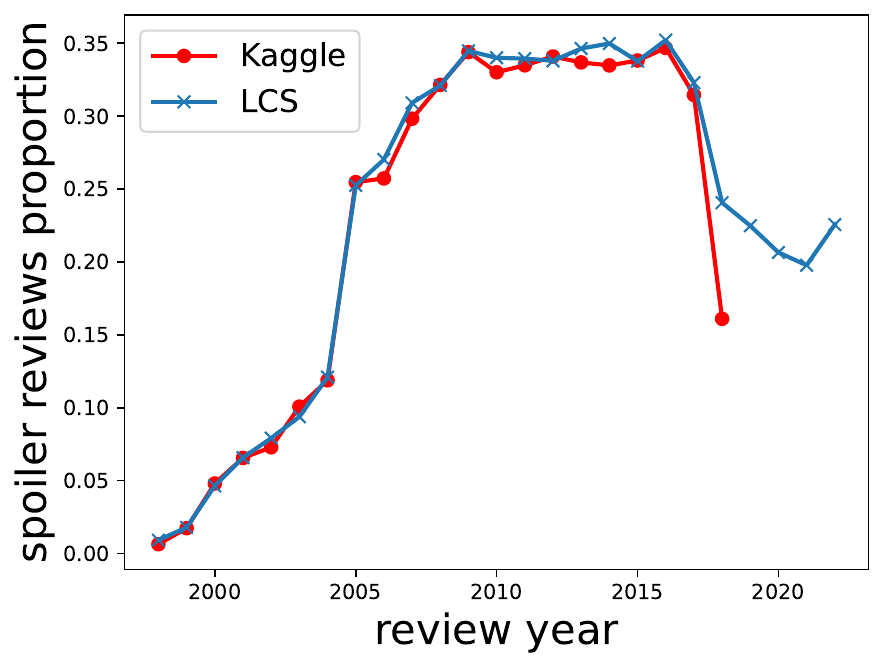}}
    \subfigure[]{
        \includegraphics[width=0.3\linewidth]{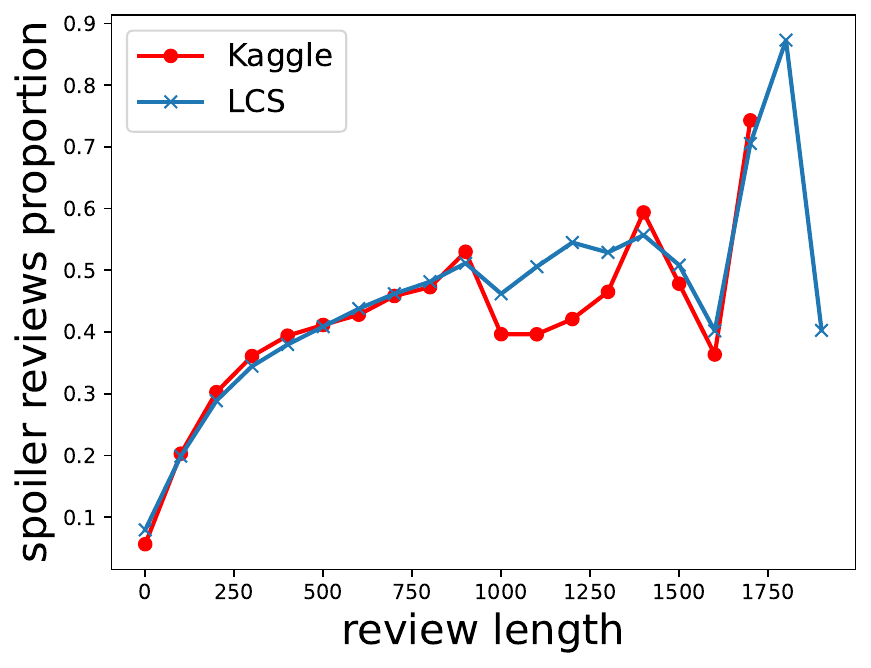}}
    \caption{(a) The spoiler proportion of reviews with different ratings; (b) The spoiler proportion of reviews posted in different years; (c) The spoiler proportion of reviews in different lengths;}
    \label{fig:metadatas}
\end{figure*}

\begin{figure}[htbp]
    \includegraphics[width=0.45\textwidth]{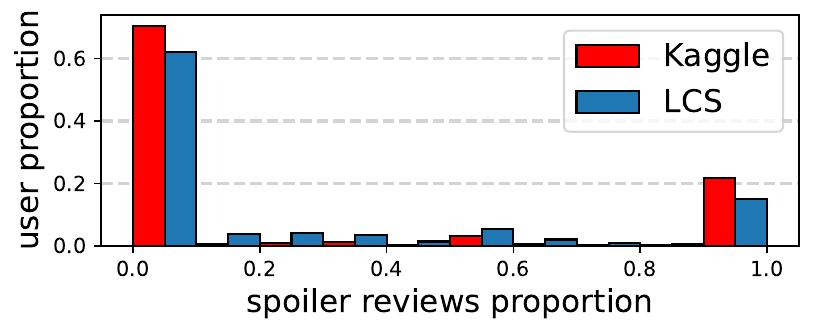}
    \caption{The proportion of spoiler reviews per user in 2 datasets, LCS and Kaggle. Spoiler review percentage intervals are divided every 10 percent.}
    \label{fig:bias}
\end{figure}

Table \ref{meta1} and Table \ref{meta2} show the metadata details of LCS and Kaggle datasets, respectively. 

We further investigate the correlation between spoilers and review ratings, publication time, and length, as depicted in Figure \ref{fig:metadatas}. Notable patterns emerge from our investigation:
\begin{itemize}[leftmargin=9pt]
    \item Spoiler reviews are often poorly rated. Highly rated reviews often reveal little about the plot.
    \item Spoiler reviews proportion in the early and recent years is low. A large number of reviews from around 2009-2016 are filled with spoilers.
    \item Longer reviews are more likely to include spoilers, suggesting that the presence of spoilers increases as the length of the review expands.
\end{itemize}

We also show the proportion of spoiler reviews per user in the 2 datasets in Figure \ref{fig:bias}. It is obvious that most users are concentrated on both ends, that is, they either barely publish spoiler reviews or publish them frequently, thus have a clear tendency.

\begin{table}[htbp]
\centering
\caption{Details of metadata contained in Kaggle.}
\resizebox{0.5\textwidth}{!}
{
\begin{tabular}{cc}
\toprule
\textbf{Entity Name} & \textbf{Metadata} \\
\midrule
User & badge count, review count, description length  \\
Movie & year, isAdult, runtime, rating, vote count, synopsis length \\
Review & time, helpful vote count, total vote count, point, content length\\
\bottomrule
\end{tabular}
}
\label{meta1}
\end{table}

\begin{table}[htbp]
\centering
\caption{Details of metadata contained in LCS.}
\resizebox{0.5\textwidth}{!}
{
\begin{tabular}{cc}
\toprule
\textbf{Entity Name} & \textbf{Metadata} \\
\midrule
Movie & year, runtime, rating, synopsis length \\
Review & time, point, content length\\
\bottomrule
\end{tabular}
}
\label{meta2}
\end{table}

\begin{table}[h]
\centering
\caption{Hyperparameter settings of \ourmethod{}.}
\resizebox{0.5\textwidth}{!}
{
\begin{tabular}{lccc}
\toprule
\textbf{Hyperparameter} & \textbf{Language Model} &
\textbf{User Transformer} & \textbf{Backbone Network} \\
\midrule
optimizer & AdamW & AdamW & AdamW \\
learning rate & 1e-5 & 1e-5 & 1e-4 \\
lr scheduler & WarmUpLinear & Exponential & Exponential \\
warm up/gamma & 0.1 & 0.9 & 0.95 \\
weight decay & 1e-3 & 1e-5 & 1e-4 \\
epochs & 1 & 20 & 60 \\
dropout & 0.1 & 0.1 & 0.2 \\
w & $\backslash$ & $\backslash$ & 1e-2 \\
\bottomrule
\end{tabular}
}
\label{hyper}
\end{table}

\begin{table}[t] \small
\centering
\caption{Model architecture parameters of \ourmethod{} on LCS dataset.}
\begin{tabular}{lc}
\toprule
\textbf{Parameters} & \textbf{Value}\\
\midrule
language model MLP hidden dim & 3072 \\
language model MLP out dim & 768 \\
\midrule
User Transformer dim & 768 \\
User Transformer feedforward dim & 3072 \\
User Transformer number of heads & 12 \\
User Transformer layers & 12 \\
User Transformer max length & 16 \\
\midrule
meta dim & 6 \\
meta MLP hidden dim & 768 \\
meta MLP out dim & 256 \\
\midrule
text projection dim & 256 \\
\midrule
GNN input dim & 774 \\
GNN hidden dim & 512 \\
GNN out dim & 256 \\
GNN layers & 2 \\
\midrule
number of experts & 2 \\
k & 1 \\
MoE MLP hidden dim & 1024 \\
MoE MLP out dim & 256 \\
\midrule
Fusion Transformer dim & 256 \\
Fusion Transformer feedforward dim & 1024 \\
Fusion Transformer number of heads & 4 \\
Fusion Transformer layers & 4 \\
\bottomrule
\end{tabular}
\label{architect1}
\end{table}

\begin{table}[h] \small
\centering
\caption{Model architecture parameters of \ourmethod{} on Kaggle dataset.}
\begin{tabular}{lc}
\toprule
\textbf{Parameters} & \textbf{Value}\\
\midrule
language model MLP hidden dim & 3072 \\
language model MLP out dim & 768 \\
\midrule
User Transformer dim & 768 \\
User Transformer feedforward dim & 3072 \\
User Transformer number of heads & 12 \\
User Transformer layers & 12 \\
User Transformer max length & 4 \\
\midrule
meta dim & 4 \\
meta MLP hidden dim & 768 \\
meta MLP out dim & 256 \\
\midrule
text projection dim & 256 \\
\midrule
GNN input dim & 772 \\
GNN hidden dim & 512 \\
GNN out dim & 256 \\
GNN layers & 2 \\
\midrule
number of experts & 4 \\
k & 1 \\
MoE MLP hidden dim & 1024 \\
MoE MLP out dim & 256 \\
\midrule
Fusion Transformer dim & 256 \\
Fusion Transformer feedforward dim & 1024 \\
Fusion Transformer number of heads & 4 \\
Fusion Transformer layers & 4 \\
\bottomrule
\end{tabular}
\label{architect2}
\end{table}

\section{Hyperparameters} \label{B}
Table \ref{hyper} illustrates the hyperparameter settings in the experiments. Table \ref{architect1} and Table \ref{architect2} demonstrate detailed model architecture parameters for easy reproduction.

\section{Experiment Details} \label{C}
\begin{itemize}[leftmargin=9pt]
    \item We use Neighbor Loader in Pytorch Geometric library to sample review nodes in the graph. We set the maximum number of neighbors to 200 and sample the 2-hop subgraph.
    \item We pad the metadata to the same dimension with -1.
    \item The Kaggle dataset doesn't provide the description of users. This situation further highlights the value of our user profile extraction module because it extracts user profiles from reviews. For GNN-based methods, we use zero vectors as the user's initial embedding. For our method \ourmethod{}, we set the first token of the sequence as learnable parameters, which is similar to the CLS token of BERT \cite{devlin2018bert}. 
\end{itemize}

\section{Baseline Details} \label{D}
We compare \ourmethod{} with PLM-based methods, GNN-based methods, and task-specific methods to ensure a holistic evaluation. For pre-trained language models, we pass the review text to the model, average all tokens, and adopt two linear projection layers to classify. For GNN-based methods, we pass the review text to RoBERTa, averaging all tokens to get the initial node feature. We provide a brief description of each of the baseline methods, in the following.
\begin{itemize}[leftmargin=9pt]
    \item \textbf{BERT} \cite{devlin2018bert} is a pre-trained language model which uses masked language model and next sentence prediction tasks to train on a large amount of natural language corpus.
    \item  \textbf{RoBERTa} \cite{liu2019roberta} is an improvement model based on BERT which removes the next sentence prediction task and improves the masking strategies.
    \item \textbf{BART} \cite{lewis2019bart} is a pre-trained language model that improves upon traditional autoregressive models by incorporating bidirectional encoding and denoising objectives. 
    \item \textbf{DeBERTa} \cite{he2021debertav3} is an advanced language model that enhances BERT by introducing disentangled attention and enhanced mask decoder.
    \item  \textbf{GCN} \cite{kipf2016semi} is a basic graph neural network that effectively captures and propagates information across graph-structured data by performing convolutions on the graph's nodes and their neighboring nodes.
    \item \textbf{R-GCN} \cite{schlichtkrull2018modeling} is an extension of GCN that specifically handles multi-relational graphs by incorporating relation-specific weights.
    \item  \textbf{GAT} \cite{velickovic2017graph} is a graph neural network that utilizes attention mechanisms to assign importance weights to neighboring nodes dynamically.
    \item  \textbf{Simple-HGN} \cite{lv2021we} is a graph neural network model designed for heterogeneous graphs, which effectively integrates multiple types of nodes and edges by employing a shared embedding space and adaptive aggregation strategies.
    \item \textbf{DNSD} \cite{chang2018deep} is a spoiler detection method using a CNN-based genre-aware attention mechanism.
    \item \textbf{SpoilerNet} \cite{wan2019fine} incorporates the hierarchical attention network (HAN) \cite{yang2016hierarchical} and the gated recurrent unit (GRU) \cite{cho2014learning} with item and user bias terms for spoiler detection.
    \item \textbf{MVSD} \cite{wang2023detecting} utilizes external movie knowledge and user networks to detect spoilers.
\end{itemize}
\end{document}